%% file: main-short.tex
\documentclass[11pt]{article}

\ifx\review\undefined
\usepackage[]{acl}
\else
\usepackage[review]{acl}
\fi

\usepackage{times}
\usepackage{hyperref}
\usepackage{amsmath}
\usepackage{amssymb}
\usepackage{geometry}
\usepackage[T1]{fontenc}
\usepackage[utf8]{inputenc}
\usepackage{inconsolata}
\usepackage{graphicx}
\usepackage[belowskip=-0.4em,labelfont=bf,labelsep=endash]{caption}
\usepackage{booktabs}
\usepackage{tabularx}
\usepackage{longtable}
\usepackage{multirow}
\usepackage{subcaption}
\usepackage{float}
\usepackage{stfloats}
\usepackage{xspace}
\usepackage[table]{xcolor}

\makeatletter
\newcommand{\sectionindicator}{\S\@gobble}
\makeatother

\definecolor{targetcolor}{HTML}{2CA02C}
\definecolor{globalcolor}{HTML}{3B7FB1}
\definecolor{localcolor}{HTML}{FF5353}
\definecolor{figonebackground}{HTML}{FDFEFF}

\definecolor{paperGreen}{HTML}{1B7837}
\definecolor{paperRed}{HTML}{8E1B1B}
\definecolor{paperSlate}{HTML}{4C566A}
\definecolor{paperGold}{HTML}{B8860B}
\definecolor{paperPurple}{HTML}{6A3D9A}
\definecolor{paperLightGreen}{HTML}{E8F3EA}
\definecolor{paperLightRed}{HTML}{F7E6E6}
\definecolor{paperLightGray}{HTML}{F2F2F2}
\definecolor{paperMidGray}{HTML}{9A9A9A}

\newcommand{\eg}{\emph{e.g.}}

\newcommand{\wpa}{\textsc{wpa}\xspace}
\newcommand{\wpafull}{\textsc{word-pair asymmetry}\xspace}

\newcommand{\shuffle}{$\Delta H$\xspace}

\author{
    \textbf{Wessel Poelman}{\normalfont\textsuperscript{1}} \quad \textbf{Kushal Tatariya}{\normalfont\textsuperscript{1}} \\
    \textbf{Jonas Mayer Martins}\textsuperscript{2} \quad \textbf{Coleman Haley}\textsuperscript{1} \quad \textbf{Miryam de Lhoneux}\textsuperscript{1}\\
 \textsuperscript{1}Department of Computer Science, KU Leuven\quad
 \textsuperscript{2}University of G\"ottingen
\\
 \small{
   \href{mailto:wessel.poelman@kuleuven.be}{\texttt{wessel.poelman@kuleuven.be}}
 }
}

\title{Measuring Word-Order Flexibility Without Syntactic Annotations}

\begin{document}
\maketitle
\begin{abstract}
    Languages exhibit different levels of flexibility regarding word order. Metrics that use syntactic annotations to measure this flexibility are limited in their applicability. We propose \wpafull, a corpus-driven metric that does not require annotations. We validate this metric against existing annotation- and typology-based metrics, finding that it (1) correlates significantly with both, and (2) correlates more strongly than an annotation-free metric inspired by previous shuffling-based approaches. We also investigate how \mbox{\wpafull} behaves across data sizes, domains, and parameters, showing that reliable estimates can be obtained with as few as a thousand sentences.
\end{abstract}

\begin{center}
    \textbf{Code:} \href{https://github.com/WPoelman/word-order-flexibility}{\texttt{word-order-flexibility}}
\end{center}

\section{Introduction}

It has been widely recognized, since at least \citet{weil-1844-question}, that some languages permit freer or more flexible word orders than others. The word-order flexibility of a language affects its learnability for both humans \citep{nowak-baggio-2017-developmental} and language models \citep[\eg,][]{kallini-etal-2024-mission,someya-etal-2025-information,mayermartins-etal-2026-vocabulary}. In natural languages, word-order flexibility and morphological richness have been argued to trade off optimally, making them learnable and efficient \citep{hockett-1958-course,gibson-etal-2019-how,bentz-etal-2023-complexity}. 

Despite its centrality to these linguistic and cognitive issues, the measurement of word-order flexibility remains a challenge in typology \cite{levshina-etal-2023-why}. Existing approaches are limited to a few order-related constructions encoded in binary typological features, use model-dependent estimates, or require syntactic annotations.

\begin{figure}[tbp]
    \centering
    \includegraphics[width=\linewidth]{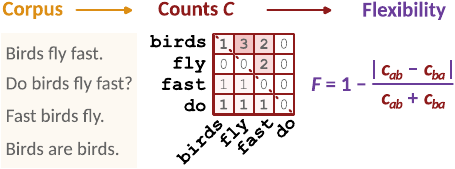}
    \caption{To measure \wpafull (\wpa), we extract from a corpus a matrix $C$ of how often word pairs co-occur in a sentence. \wpa is a measure of how symmetric this matrix is, \eg, whether \texttt{birds} always occurs before \texttt{fly}.}
    \label{fig:1}
\end{figure}

Can a measure of word-order flexibility be corpus-driven, model-independent, and remain independent of any particular syntactic analysis? We introduce \wpafull (\wpa), a flexibility metric that requires no syntactic annotations, making it applicable to any tokenized corpus. We validate \wpa against annotation-based metrics from \citet{futrell-etal-2015-quantifying} and language-level ratings from \citet{bakker-1998-flexibility}. We find that \wpa correlates significantly and positively with both, contrary to a shuffling-based metric inspired by \citet{montemurro-zanette-2011-universal}. In addition, we examine \wpa across weighting strategies, corpus sizes, and domains to test its robustness.

\section{Related Work}\label{sec:related-work}
This section reviews existing approaches to measuring word-order flexibility, highlighting the limitations our study aims to address.

\paragraph{Typological features.}

Some studies measure word-order flexibility by combining several binary typological features (\eg, whether adjectives appear both before and after nouns) into an aggregate score  \citep{bakker-1998-flexibility}. However, these approaches necessarily focus only on a few select types of relations between words, potentially missing variation (or a lack thereof) in other constructions as well as frequency and distributional properties---they cover only a limited subset of word-order properties in a language.

\paragraph{Annotation-based.}
Most corpus-based measures of word-order flexibility use syntactic parses, often from Universal Dependencies \cite[UD;][]{nivre-etal-2020-universal}. Several works use entropy-based measures that look at the distribution of heads and dependents \cite{futrell-etal-2015-quantifying,gerdes-etal-2019-rediscovering,kahane-etal-2023-word}. Others use proportions of certain constructions and whether the order in which they occur is balanced or not \cite{baylor-etal-2024-multilingual}. We discuss the approach of \citet{futrell-etal-2015-quantifying}, the first and most influential, in more detail in \autoref{sec:metrics}.

Although these approaches account for all relations between words in a corpus, the syntactic annotations they rely on are both expensive to obtain and theory-dependent. This limits their applicability to languages with annotated corpora and restricts related findings to their respective syntactic formalisms.

\paragraph{Shuffling-based.} \citet{montemurro-zanette-2011-universal} define the flexibility of word-order in a language as the difference in word entropy between the language itself and a shuffled variant. The signal that remains is then the information due to word order. This measure relies on unbiased entropy estimators, which may not hold for current language models (see \autoref{apx:montemurro}). This prior work has also folded in cross-sentence discourse and information structure, obscuring a purely syntactic interpretation, a complication that we address \cite[\textit{cf.}][]{koplenig-etal-2017-statistical}.

\section{Method}\label{sec:method}

In this section, we define \wpa, comparable metrics, and our experimental setup.

\subsection{Word-Pair Asymmetry}\label{sec:wpa}
We introduce \wpafull (\wpa), a measure of word-order flexibility based on all pairs of words in a sentence: if for words $a$ and $b$, both orders $(a,b)$ and $(b,a)$ occur equally, this pair is maximally flexible, see \autoref{fig:1}. Conversely, if only one order occurs, the pair is maximally rigid.
Pair frequencies are also weighted by the distance between the two words.
Adjacent pairs tend to reflect common structural patterns, such as those in dependency annotations (determiner-noun, adjective-noun, auxiliary-verb), while distant pairs in a sentence share less mutual information \cite{futrell-etal-2015-largescale}. 
Concretely, we weight the frequencies with an exponential distance decay such that if $a$ occurs before $b$ at distance $d = k - j$, the observation is weighted with $\lambda^d$, with $0<\lambda\le1$. Setting \(\lambda=1\) gives equal weight to all pairs; smaller values of $\lambda$ prioritize local constructions.

Formally, $\mathcal{C}=\{S_i\}_{i=1}^{N}$ is a corpus of sentences $S_i=(w_{i,1},\ldots,w_{i,n_i})$.\footnote{We exclude punctuation here. For the UD-based metrics, we additionally omit \texttt{punct} dependency relations and tokens headed by \texttt{PUNCT}, following \citet{futrell-etal-2015-quantifying}.}
The distance-weighted count of $a$ occurring before $b$ is
\begin{equation}
\begin{aligned}
  c^{(\lambda)}_{ab}
  &=\sum_{i=1}^N \; \sum_{1\le j<k\le n_i}
    \lambda^{k-j} \\
  &\quad\cdot
    \mathbf{1}[w_{i,j}=a]\mathbf{1}[w_{i,k}=b].
\end{aligned}
\end{equation}
The pairwise \emph{flexibility} is the balance of orders:
\begin{equation}
  F^{(\lambda)}_{ab}
    =1-\frac{|c^{(\lambda)}_{ab}-c^{(\lambda)}_{ba}|}
            {c^{(\lambda)}_{ab}+c^{(\lambda)}_{ba}}\,,
\end{equation}
which is 0 when all observations favor one order and 1 when both orders are equally frequent.
We invert the asymmetry to make the final score more intuitive: larger values indicate more flexibility.

The flexibility on the corpus-level is the mean over qualifying pairs $\mathcal{P}$,
\begin{equation}\label{eq:aggregate}
  F^{(\lambda)}
    =\frac{1}{|\mathcal{P}|}
      \sum_{\{a,b\}\in\mathcal{P}}F^{(\lambda)}_{ab}.
\end{equation}
In other words, $F^{(\lambda)}$ measures how symmetric the co-occurrence matrix $C = \{c_{ab}\}$ is. We only consider pairs that occur at least three times across both directions combined (unweighted).
This final aggregate at the corpus-level is what we refer to as \wpa or, to make the rate $\lambda$ explicit, $F^{(\lambda)}$.

\subsection{Comparison to existing metrics}\label{sec:metrics}
We compare \wpa to a representative metric from each of the three approaches mentioned in \autoref{sec:related-work}: typological ratings, annotations, and shuffling.
For the typological ratings, we use the \emph{indices} defined by \citet{bakker-1998-flexibility}.
These scores summarize how often a language allows more than one order for a given construction.

\paragraph{Annotation-based.} 
For measures built on syntactic annotations, we rely on UD-based metrics by \citet{futrell-etal-2015-quantifying}, who introduce \emph{Head-Direction Entropy} (HDE) and \emph{Relation-Order Entropy} (ROE).

HDE measures the entropy $H$ of dependency directions, based on the relation type and the part-of-speech tags of the head and dependent.\footnote{To estimate $H$, \citeauthor{futrell-etal-2015-quantifying} use the bootstrap correction from \citet{dedeo-etal-2013-bootstrap}.
We instead use the \citet{chao-shen-2003-nonparametric} entropy estimator, which \citet{arora-etal-2022-estimating} recommend for sparse linguistic data.}
Formally, the dependency direction \(D\in\{l,r\}\) can be left $l$ or right $r$, depending on the context
$K =(\mathrm{deprel},\mathrm{upos}_{\mathrm{head}},\mathrm{upos}_{\mathrm{dep}})$, \emph{i.e.}, the type of dependency relation and the universal part-of-speech tags of the head and its dependent, such that
$\operatorname{HDE} = H(D\mid K)\,.$

ROE measures whether a given local dependency structure is consistently realized in one (rigid) or several (flexible) linear orders.
Letting~$U$ denote an unordered local subtree and $L$ its linearization,
$\operatorname{ROE} = H(L\mid U).$

\paragraph{Shuffling-based.}
The closest prior work on annotation-free estimation of flexibility is by \citet{montemurro-zanette-2011-universal}.
Their metric is based on the intuition that the information due to the word order in a language is given by the difference in entropy between the language itself and a shuffled version of that language without a fixed word order. 
However, \citeauthor{montemurro-zanette-2011-universal} shuffle words across entire documents, thus conflating sentence-internal order and the order of sentences and paragraphs. To isolate the information from sentence-internal order, we permute words within each sentence instead.

To estimate the word entropy, we run a trial~$r \in \{1,\dotsc,R\}$, where we draw a permutation~$\pi_r$ of the sentence indices $\{1,\dotsc,N\}$ uniformly at random to concatenate the sentences~$S$ in that order. This produces a reference text
\begin{equation}
    X_r = S_{\pi_r(1)}\mathbin{\|}S_{\pi_r(2)}\mathbin{\|}\dotsb \mathbin{\|}S_{\pi_r(N)}\,.
\end{equation}
Independently for each sentence $i$, we also sample a permutation $\sigma_{r,i}$ of the word indices $\{1,\dots,n_i\}$ uniformly at random, writing the sentence with shuffled words as $\sigma(S_i)=(w_{i,\sigma(1)},\dots,w_{i,\sigma(n_i)})$.
We then construct our shuffled text as
\begin{align}
    Y_r &=
    \sigma_{r,\pi_r(1)} \big(S_{\pi_r(1)}\big) \mathbin{\|} \sigma_{r,\pi_r(2)} \big(S_{\pi_r(2)}\big) \mathbin{\|}  \notag\\
    &\quad \dotsb \mathbin{\|} \sigma_{r,\pi_r(N)} \big(S_{\pi_r(N)}\big)\,.
\end{align}
The two texts $X_r$ and $Y_r$ thus have the same sentence order but the latter has a shuffled sentence-internal word order.

We then follow \citeauthor{montemurro-zanette-2011-universal} and use the Lempel--Ziv compression algorithm \citep{ziv-lempel-1977-universal} as an entropy estimator for \emph{both texts} (as there is no longer an analytic entropy estimate for the shuffled text), and average the difference between the entropy of $X_r$ and $Y_r$ across our $R$ trials.\footnote{Note that we reverse the sign from \citeauthor{montemurro-zanette-2011-universal} to match the logic of the other metrics: larger values correspond to \emph{greater} word-order flexibility (and \emph{less} information due to the word order).} Denoting the estimated entropy of a text as $\widehat{H}(\cdot)$ in bits per token, the entropy difference is thus
\begin{equation}
    \Delta H = \frac{1}{R} \sum_{r=1}^{R}
    \big[
        \widehat{H}(X_r) - \widehat{H}(Y_r)
    \big]\,.
\end{equation}

\paragraph{Controlling for lexical diversity.}
Corpora with a high \emph{type-token ratio} (TTR) yield more distinct word pairs and therefore sparser distributions, which could inflate the data-driven metrics (not the indices by \citeauthor{bakker-1998-flexibility}).
Therefore, we filter out rare pairs for $F^{(\lambda)}$ (\autoref{eq:aggregate}), and report raw and TTR-controlled correlations.
These correlations show whether metrics are related beyond depending on lexical diversity.
With HDE, for example, the TTR-controlled Pearson correlation is
\begin{equation}
  r(F,\operatorname{HDE}\mid\operatorname{TTR})
  =r(F-\widehat F,\;\operatorname{HDE}-\widehat{\operatorname{HDE}})\,,
\end{equation}
where $\widehat F$ and $\widehat{\operatorname{HDE}}$ are predictions from separate linear regressions on the TTR. 
Spearman rank correlations are computed on the same residuals.

\paragraph{Experimental setup.}
We use data from UD~v2.17 \cite{zeman-etal-2025-universal} for the comparison between \wpa, HDE, ROE, and \shuffle.
First, we use exactly the same sentences to calculate the metrics ($1{,}000$ per language, following \citeauthor{futrell-etal-2015-quantifying}).
Second, we compare all metrics to the ratings from \citet{bakker-1998-flexibility}.
Third, we perform ablation studies based on the choice of the rate~$\lambda$ and across corpus sizes.
Lastly, we apply \wpa and \shuffle to raw-text corpora to see how stable they are within and across domains.\footnote{See \autoref{tab:language-coverage} for language coverage per analysis.}

\section{Results}\label{sec:results}
\begin{figure}[t]
    \centering
    \includegraphics[width=\linewidth]{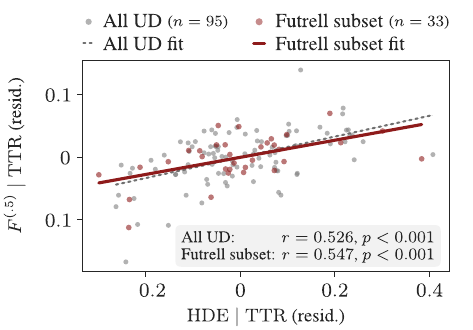}
    \caption{Validation of \wpa with HDE. We select languages that have at least $1{,}000$ sentences.}
    \label{fig:2}
\end{figure}

\paragraph{Validation.}
We compare the \wpa scores against HDE and ROE for 33 of the 34 languages used by \citeauthor{futrell-etal-2015-quantifying},\footnote{We exclude Bengali since it has $< 1{,}000$ sentences.} and a much larger sample of 95 languages. 
In both settings, there is a significant positive correlation between \wpa and HDE at $\lambda=.5$ (see \autoref{fig:2}). Other values for~$\lambda$ are discussed below. With ROE, the correlation is much less clear (see \autoref{fig:8a}) even when controlling for TTR. This is in part due to the sensitivity of ROE to data size, and due to many languages having almost identical values for ROE.\footnote{See Figure 2, page 9 in \citet{futrell-etal-2015-quantifying}.}
The correlation is very similar on the 95 languages, again with $1{,}000$ sentences per language.
We show more detailed analyses regarding the effect of controlling for TTR in \autoref{apx:stability}, \autoref{fig:additional-results}.

\begin{figure}[t]
    \centering
    \includegraphics[width=\linewidth]{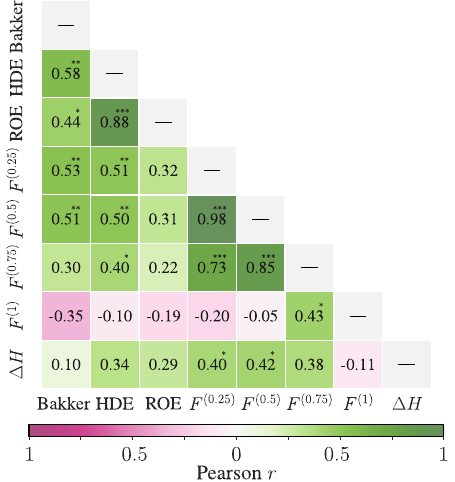}
    \caption{Correlation heatmap between annotation-based metrics and \wpa at \mbox{$\lambda \in \{.25, .5, .75, 1\}$}. We use the intersection of 33 languages with \citet{bakker-1998-flexibility}, see \autoref{tab:language-coverage}. Note that these values are different from \autoref{fig:2} since we do not control for TTR given that \citeauthor{bakker-1998-flexibility}'s ratings are on the language level. We apply the FDR correction from \citet{benjamini-hochberg-1995-controlling} on the lower triangle (excluding the diagonal): $^{*}q<.05$; $^{**}q<.01$; $^{***}q<.001$.}
    \label{fig:3}
\end{figure}

\autoref{fig:3} shows a correlation grid of the comparison metrics and \wpa. HDE correlates moderately with the \citeauthor{bakker-1998-flexibility} ratings ($r=.58$), while the ROE--\citeauthor{bakker-1998-flexibility} correlation is weaker ($r=.44$). \wpa at $\lambda=.25$ is very similar to HDE in its correlation with the \citeauthor{bakker-1998-flexibility} ratings  ($r=.53$). This suggests that \wpa captures a signal similar to both language- and UD-based flexibility metrics.

We observe fairly weak, mostly non-significant correlations between \shuffle and all other metrics. We do observe significant correlations with \wpa for $\lambda=.25$ and $\lambda=.5$, suggesting that both metrics may pick up on shared aspects of the text corpora they are estimated on.

\paragraph{Ablating $\boldsymbol{\lambda}$.}
From \autoref{fig:3}, we observe that a larger~$\lambda$ leads to weaker correlations between \wpa and existing measures. At $\lambda=1$ (\emph{i.e.,} when distance is ignored and pairs are weighted equally), most correlations are even negative. A~potential explanation for this is that both at the UD-level and at the typological level, local ordering is more important. Most dependents are close (\eg, determiner-noun) and most typological ratings are also based on close relations (\eg, adjective-noun). Put differently, at $\lambda=1$, \wpa is essentially a ``flat'' and pairwise view of a sentence where distance and locality are ignored.

\begin{figure}[t]
    \centering
    \includegraphics[width=\linewidth]{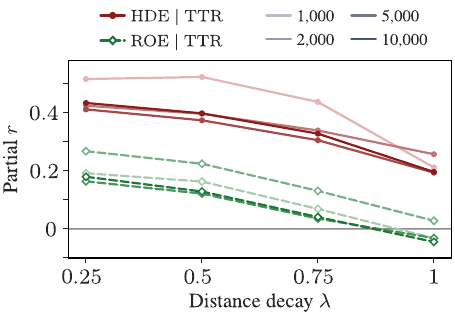}
    \caption{Effect of corpus size and $\lambda$ on correlations. This is calculated at each size using the same 32 languages that have at least $10{,}000$ sentences.}
    \label{fig:4}
\end{figure}

\paragraph{Ablating corpus size.}
We study the effect of corpus size in \autoref{fig:4} with a fixed set of languages that have at least $10{,}000$ sentences available ($n=32$). In terms of choosing the rate~$\lambda$, the correlation with HDE and ROE is robust across all sizes, indicated by small vertical differences. Thus, $1{,}000$ sentences appear to be a reasonable sample size for obtaining reliable estimates, both for the \citeauthor{futrell-etal-2015-quantifying} metrics and \wpa.\footnote{Also, see \autoref{fig:8c} in \autoref{apx:stability}.}

\begin{figure}[t]
    \centering
    \includegraphics[width=\linewidth]{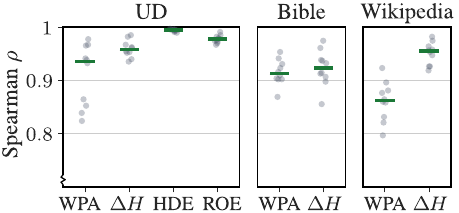}
    \caption{Spearman rank correlation for robustness within domains. For the same 20 languages, we draw five disjoint samples of $1{,}000$ sentences per domain. The plot shows whether the resulting ranks are stable across samples and languages. We use $\lambda = .5$. The green line is the median of the ten pairwise correlations.}
    \label{fig:5}
\end{figure}

\paragraph{Ablating domain.}
Since \wpa does not require annotations, it can be calculated on any corpus. We choose Wikipedia and the Bible as domains comparable to UD. \autoref{fig:5} shows whether the metrics are consistent \emph{within} domains. Five disjoint sets of $1{,}000$ sentences are sampled per domain per language to test if the ranking of languages is consistent. We observe that both \wpa and \shuffle are quite robust within the same domain. HDE and ROE are even more stable, but this is to be expected since UD's dependency and part-of-speech annotations abstract away lexical diversity, while \wpa and \shuffle do not.

Using the same setup as in \autoref{fig:5}, we also compare \emph{across} domains in \autoref{tab:cross-domain}. Across domains, robustness is much harder to achieve, given that \wpa also captures lexical diversity (pair \emph{instances}, not just relations).

\begin{table}[H]
    \centering
    \small
    \input{tab/cross-domain}
    \caption{Spearman rank correlation for robustness across domains for 20~languages, after averaging the five sample scores per language and domain. $^{*}$BH-adjusted $q < .05$.}
    \label{tab:cross-domain}
\end{table}

These results show that flexibility \emph{within} domains can be measured quite robustly with a small number of sentences, but consistent flexibility \emph{across} domains may be difficult to obtain with a corpus-driven metric. There is a difference between characterizing the flexibility of a language as a whole versus \emph{specific corpora}.  

\section{Conclusion}
We present \wpafull (\wpa), an annotation-free measure of word-order flexibility. \wpa computes how balanced the two possible orders of word pairs are, weighting observations by their distance exponentially through a rate~$\lambda$. Our results show moderate and significant correlations with \emph{Head-Direction Entropy}, a UD-based metric proposed by \citet{futrell-etal-2015-quantifying}, as well as with typological flexibility ratings from \citet{bakker-1998-flexibility}. We also compare \wpa with an annotation-free metric inspired by \citet{montemurro-zanette-2011-universal} and based on word-order entropy, which does not itself correlate significantly with either \citeauthor{futrell-etal-2015-quantifying} or \citeauthor{bakker-1998-flexibility}. Lastly, we test \wpa across different decay rates~$\lambda$ and corpus sizes, and find that (1) local dependencies are the most important, (2) the metric is robust across corpus sizes and within domains, and (3) reliable estimates can be obtained with $1{,}000$ sentences.

The corpus-grounded, annotation-free, and unit-agnostic measure of \wpa can enable new types of studies, expanding our understanding of both linguistic diversity and NLP systems. Because \wpa is corpus-based, it could be used to study variation in the word-order flexibility of a language across domains. The annotation-free nature of \wpa enables data-driven studies of word order in languages that do not have syntactically annotated corpora available. It therefore makes large-scale corpus-based typological studies \cite[\eg,][]{levshina-etal-2023-why} possible, for instance, a study of how syntactic complexity trades off with other dimensions of linguistic complexity like morphology. Because it provides a measure of syntactic complexity that can readily be extended to counterfactual languages, it could be used to formulate hypotheses about learnability constraints for both humans and language models. Further, because \wpa can be measured using any tokenized corpus, it could straightforwardly be applied at different levels of granularity (\eg, at the morpheme- or subword-token level) to investigate how different units of analysis affect apparent order variability. This could be useful for understanding how typology affects language model learnability under subword tokenization \cite[\eg,][]{tatariya-etal-2025-interplay}. The same property also makes it easily adaptable to languages where conventional \textit{word} segmentation is difficult to obtain or not applicable.

\clearpage

\section*{Limitations}

Any study of word-order flexibility necessarily relies on some notion of a \textit{word}, and there is no uncontroversial operationalization \cite{haspelmath-2011-indeterminacy,haspelmath-2023-defining,mansfield-2021-word}. In this study, we follow prior work in relying on the word boundaries defined in the Universal Dependencies project and by the machine translation tokenizers from \citet{post-2018-call}, but these are not without contention, and the choice of word segmentation could affect the rankings of languages for all metrics discussed here.

Moreover, while \wpa correlates with existing, more linguistically grounded metrics of word-order flexibility, it does not explicitly account for linguistic differences in observed variations of word order and polysemy, which is an inherent property of corpus-driven word-level statistics.

Finally, although we aim to disentangle morphology from word order by controlling for TTR and filtering out rare pairs, we cannot fully account for all factors to separate these two intricately related phenomena \cite{bentz-etal-2023-complexity}.

\ifx\review\undefined
\section*{Acknowledgments}
We thank the anonymous reviewers for their comments.
WP is funded by a KU Leuven Bijzonder Onderzoeksfonds C1 project with reference C14/23/096. CH is funded by the Research Foundation -- Flanders (FWO) under grant number G047025N.
The computational resources and services used were provided by the VSC (Flemish Supercomputer Center), funded by the Research Foundation -- Flanders (FWO) and the Flemish Government -- department EWI.
\fi

\section*{Use of AI}
We used AI assistance for coding, brainstorming, and final grammar checking. All output has been verified and reviewed by the authors, who also take full responsibility.

\bibliography{refs}
\clearpage

\appendix
\section{Choice of entropy estimator for \texorpdfstring{\shuffle}{word-order entropy}}\label{apx:montemurro}

We follow \citet{montemurro-zanette-2011-universal} in using a Lempel--Ziv estimator to compute the \shuffle metric. This section briefly highlights the considerations in choosing an entropy estimator, how it affects the metric design, and provides an analysis of convergence of \shuffle.

In NLP, the average entropy of a sentence in a corpus is commonly estimated with a language model. Unlike a Lempel--Ziv estimator, language models estimate entropy at the level of next word/token predictions, which means they can easily assign entropy to a set of sentences. However, traditional language models are poor candidates for estimating entropy in this context.
A transformer LM, for example, could fail to converge on the tiny corpora we have used in our study. Furthermore, transformer language models (currently the state-of-the-art approach) have recently been shown to have information-locality biases \citep{kallini-etal-2024-mission,someya-etal-2025-information,mayermartins-etal-2026-vocabulary}, such that a language with more long-distance dependencies (such as a shuffled language) yields higher entropy estimates from transformer language models, \textit{even when it has the exact same global entropy as a natural language}. While it is not known if other neural models share these biases, they may also fail to converge, making them a poor fit for the present study. The most common statistical language-model alternative, $n$-gram models, only model dependencies up to a fixed length, and incur substantial sparsity penalties for long-range dependencies, making them a poor choice for estimating syntactic ordering on small-scale data.

\begin{figure*}[b]
    \centering
    \includegraphics[width=\linewidth]{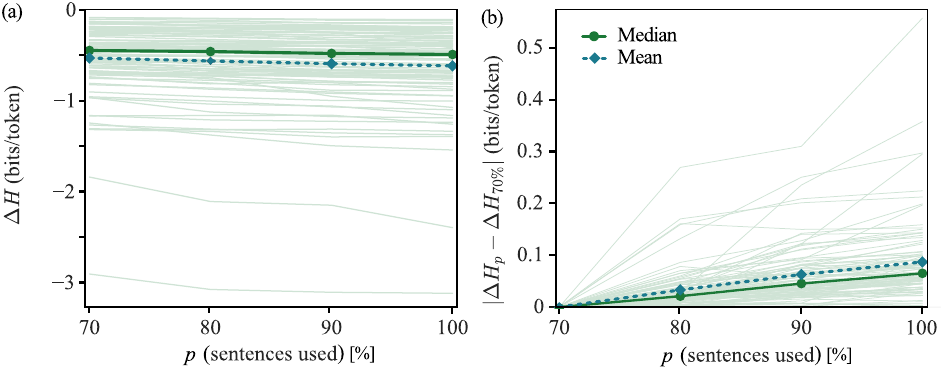}
    \caption{\shuffle grows gradually as the amount of data increases from $\sim700$ to $\sim1{,}000$ sentences, suggesting the estimate has not yet converged. (a) Entropy difference~$\Delta H$. (b) Absolute change from $70\,\%$ of the data.}\label{fig:6}
\end{figure*}

\begin{figure*}[b]
    \centering
    \includegraphics[width=\linewidth]{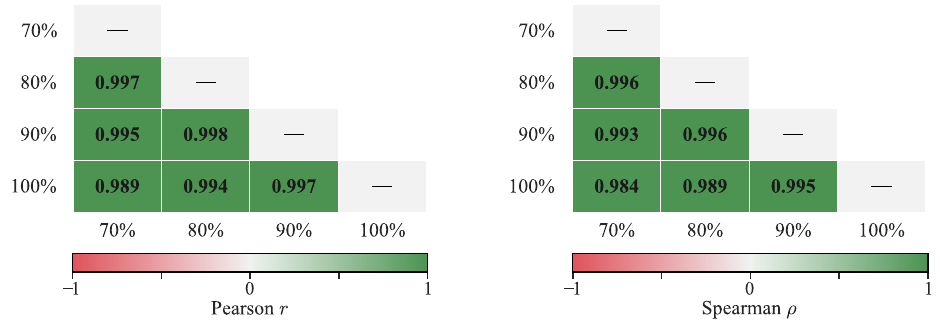}
    \caption{\shuffle is consistent in the relative values of languages as the amount of data increases, suggesting a systematic bias in entropy estimates which does not affect the relative ranking of languages (and therefore the relationship with syntactic flexibility measures).}\label{fig:7}
\end{figure*}

In contrast to the approaches with language models, the Lempel-Ziv estimator provides a token-level estimate of entropy \textit{for the average token in the corpus it is fit to}, rather than predictively for next tokens in a sequence. As a result, we fit it $R$ times to different sentence-level random shuffles of our corpus, to reduce the influence of sentence order information. This gives it more data on which to converge. To check convergence, we recompute \shuffle on 70\%, 80\%, and 90\% of our data, and compare the results to the metric computed on the whole corpus. As shown in \autoref{fig:6}, we see a gradual, small linear increase in the values of \shuffle as the number of included sentences increases, suggesting the metric has not converged to a fixed value. However, the \textit{relative} values of individual languages seem to have converged: holding out data in the same way, \autoref{fig:7} shows high Pearson and Spearman correlations between the scores of individual languages at different data levels. This suggests that the metric growth is roughly equal across languages, keeping the ranking of languages (and thus relation to other metrics) consistent. Accordingly, and given the issues with other entropy estimators in this context, we conclude this metric is not as promising as \wpa for annotation-free estimation of syntactic information.

\section{Data Preprocessing}
For the corpus-size ablations and for languages that do not have one treebank with $1{,}000$ sentences, we combine multiple treebanks for a single language. We exclude treebanks that are created from historical, dialect-specific, spoken, and poetic data. Further, we exclude code-switching and copyright-protected treebanks. For Chinese, we do not combine treebanks in traditional (\texttt{Hant}) and simplified (\texttt{Hans}) Chinese scripts together, and treat them as separate. Norwegian has two treebanks---one for Norwegian Nynorsk (\texttt{nno}) and one for Norwegian Bokm\r{a}l (\texttt{nob}). We treat these two varieties as separate languages since they have different ISO 639-3 codes. Finally, for Sanskrit there are two treebanks---one in Vedic Sanskrit and one in Classical Sanskrit. We treat these as separate since there is significant morphological and syntactic differentiation between the two varieties. Consequently, Classical Sanskrit is not included in the analysis since, on its own, the treebank has fewer than $1{,}000$ sentences. When making comparisons across domains, we only compare languages that are represented by the same script in both UD and the target dataset (the Bible and Wikipedia). For example, the correlation between the Bible and UD for Hindi is only considered for the \wpa score on \texttt{hin\_Deva} and not for \texttt{hin\_Latn} since the UD treebank is in Devanagari script. Language codes are resolved with QQ \cite{poelman-etal-2026-qq}.

For Wikipedia, we sample $5{,}000$ articles from each edition after applying the filtering strategies from \citet{tatariya-etal-2026-how}. These articles are then split into sentences using a multilingual sentence segmentation model using \texttt{WTPSplit} from \citet{frohmann-etal-2024-segment}. Sentences are tokenized into words using tokenizers from \texttt{SacreBLEU} \cite{post-2018-call}. Sentences with fewer than five words are discarded. For the Bible, we use the \emph{Parallel Bible Corpus} \cite[PBC;][]{mayer-cysouw-2014-creating}, which is already segmented into verses ($\sim$sentences) and words.
The PBC is not openly available but can be requested from the original authors.
\autoref{tab:datasets-and-software} lists datasets and software.

\begin{table}[htbp]
    \centering
    \small
    \resizebox{\linewidth}{!}{\input{tab/datasets-and-software}}
    \caption{Datasets and software used.}
    \label{tab:datasets-and-software}
\end{table}
\hfill
\clearpage

\section{Language Coverage}\label{apx-dataset-details}
See \autoref{tab:language-coverage}.

\begin{table*}[htbp]
\centering
\small
\setlength{\tabcolsep}{4pt}
\input{tab/languages}
\caption{Language sets used in each analysis.}\label{tab:language-coverage}
\end{table*}

\section{Additional Results}\label{apx:stability}
\begin{figure*}[b]
    \begin{subfigure}[t]{0.48\textwidth}
        \includegraphics[width=\linewidth]{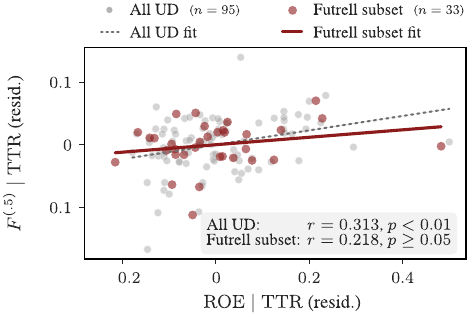}
        \caption{Correlation of \wpa with ROE from \citet{futrell-etal-2015-quantifying}.}
        \label{fig:8a}
    \end{subfigure}
    \hfill
    \begin{subfigure}[t]{0.48\textwidth}
        \centering
        \includegraphics[width=\linewidth]{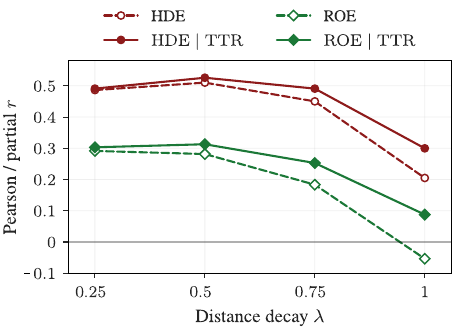}
        \caption{Effect of choosing $\lambda$. This is tested on $1{,}000$ sentences using 95 languages.}
        \label{fig:8b}
    \end{subfigure}
    \par\vspace{10pt}
    \begin{subfigure}[t]{0.48\textwidth}
        \centering
        \includegraphics[width=\linewidth]{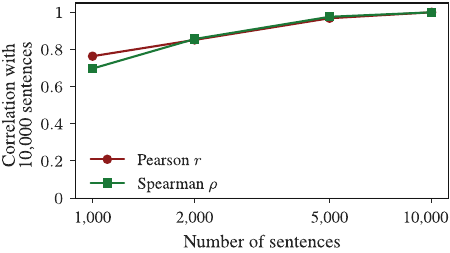}
        \caption{Stability of the rankings of languages across sentence sizes. The same 32 languages are used across all sizes; selection depends on languages having at least $10{,}000$ sentences available.}
        \label{fig:8c}
    \end{subfigure}
    \hfill
    \begin{subfigure}[t]{0.48\textwidth}
        \centering
        \includegraphics[width=\linewidth]{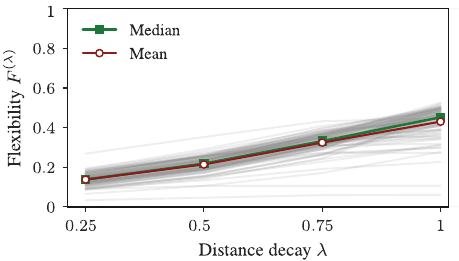}
        \caption{Behavior of \wpa for different $\lambda$ values across languages. This is for 95 languages and tested with $1{,}000$ sentences.}
        \label{fig:8d}
    \end{subfigure}
    \centering
    \begin{subfigure}[t]{0.48\textwidth}
        \includegraphics[width=\linewidth]{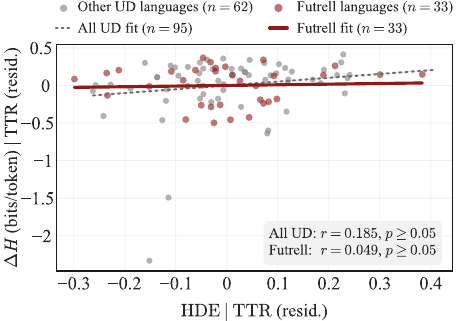}
        \caption{Validation of \shuffle with HDE. We select languages that have at least $1{,}000$ sentences, and control for type-token ratio (TTR).}
        \label{fig:8e}
    \end{subfigure}
    \caption{Additional analyses.}\label{fig:additional-results}
\end{figure*}

\autoref{fig:additional-results} shows additional analyses.
\autoref{fig:8a} shows the correlation with ROE, as discussed in \autoref{sec:results}.
\autoref{fig:8b} shows the effect of $\lambda$ on the correlations with HDE and ROE, as well as the impact of TTR on these correlations.
We can see that $\lambda=.25$ and $\lambda=.5$ are very close while $\lambda = 1$ does not capture the same information.
This also seems to confirm that nearby word pairs carry most of the signal.
\autoref{fig:8c} shows how stable the rankings are across increasing corpus sizes.
\autoref{fig:8d} shows how $F$ changes for different rates~$\lambda$.
\autoref{fig:8e} shows the detailed correlation between \shuffle and HDE.
Lastly, \autoref{tab:full-results} shows the full validation results from \autoref{sec:results}.

\onecolumn

\begin{center}
    \input{tab/full-results}
\end{center}

\end{document}

%% file: tab/cross-domain.tex
\begin{tabular}{lll}
\toprule
Domain pair & \wpa $\rho$ & $\Delta H$ $\rho$ \\
\midrule
UD--Wikipedia    & $.54^{*}$ & $.33$ \\
UD--Bible        & $.32$     & $.41$ \\
Wikipedia--Bible & $.27$     & $.43$ \\
\bottomrule
\end{tabular}

%% file: tab/datasets-and-software.tex
\begin{tabularx}{\columnwidth}{@{}l c X@{}}
    \toprule
    Resource & Version & Citation \\
    \midrule
    UD          & 2.17  & \citet{zeman-etal-2025-universal} \\
    PBC         & ---   & \citet{mayer-cysouw-2014-creating} \\
    WikiQuality & ---   & \citet{tatariya-etal-2026-how} \\
    \midrule
    Qwanqwa (QQ) & 1.2.0 & \citet{poelman-etal-2026-qq} \\
    WTPSplit     & 2.2.1 & \citet{frohmann-etal-2024-segment} \\
    SacreBLEU    & 2.6.0 & \citet{post-2018-call} \\
    \bottomrule
\end{tabularx}

%% file: tab/languages.tex
\begin{tabularx}{\textwidth}{@{}lr>{\raggedright\arraybackslash}X@{}}
\toprule
Analysis & $n$ & Languages  \\
\midrule
UD $1{,}000$ & 95 & Abkhaz, Afrikaans, Akkadian, Alemannic, Amharic, Ancient Greek, Ancient Hebrew, Arabic, Bambara, Basque, Belarusian, Bororo, Bulgarian, Cantonese, Catalan, Chinese Simplified, Chinese Traditional, Chintang, Chukchi, Coptic, Croatian, Czech, Danish, Dutch, Eastern Armenian, Egyptian, English, Erzya, Estonian, Faroese, Finnish, French, Galician, Georgian, German, Gothic, Guajajara, Haitian Creole, Hausa, Hebrew, Highland Puebla Nahuatl, Hindi, Hungarian, Icelandic, Indonesian, Irish, Italian, Japanese, Javanese, Kazakh, Khoekhoe, K'iche', Korean, Kyrgyz, Latin, Latvian, Lithuanian, Maltese, Manx, Modern Greek, Naija, Nheengatu, North Sami, Norwegian Bokm\r{a}l, Norwegian Nynorsk, Occitan, Ottoman Turkish, Persian, Polish, Pomak, Portuguese, Romanian, Russian, Vedic Sanskrit, Scottish Gaelic, Serbian, Sindhi, Slovak, Slovenian, Spanish, Swedish, Tamil, Telugu, Thai, Turkish, Ukrainian, Urdu, Uyghur, Uzbek, Vietnamese, Welsh, Western Armenian, Wolof, Yiddish, Zaar \\
UD $10{,}000$ & 32 & Ancient Greek, Belarusian, Bororo, Bulgarian, Catalan, Czech, Dutch, English, Estonian, Finnish, French, German, Hebrew, Hindi, Italian, Korean, Latin, Latvian, Norwegian Bokm\r{a}l, Norwegian Nynorsk, Persian, Polish, Portuguese, Romanian, Russian, Vedic Sanskrit, Slovak, Slovenian, Spanish, Swedish, Turkish, Ukrainian \\
\citeauthor{futrell-etal-2015-quantifying} overlap & 33 & Ancient Greek, Arabic, Basque, Bulgarian, Catalan, Croatian, Czech, Danish, Dutch, English, Estonian, Finnish, French, German, Hebrew, Hindi, Hungarian, Irish, Italian, Japanese, Latin, Modern Greek, Persian, Portuguese, Romanian, Russian, Slovak, Slovenian, Spanish, Swedish, Tamil, Telugu, Turkish \\
\citeauthor{bakker-1998-flexibility} overlap & 33 & Abkhaz, Basque, Bulgarian, Catalan, Danish, Dutch, English, Estonian, Faroese, Finnish, French, Galician, Georgian, German, Gothic, Icelandic, Irish, Italian, Latin, Latvian, Lithuanian, Maltese, Manx, Modern Greek, Polish, Portuguese, Russian, Scottish Gaelic, Slovak, Spanish, Swedish, Turkish, Yiddish \\
Domain robustness & 20 & Arabic, Basque, Czech, Dutch, German, Hebrew, Hindi, Indonesian, Italian, Korean, Naija, Polish, Portuguese, Romanian, Russian, Spanish, Swedish, Turkish, Ukrainian, Urdu \\
\bottomrule
\end{tabularx}

%% file: tab/full-results.tex
\begingroup
\footnotesize
\setlength{\tabcolsep}{3pt}
\setlength{\LTleft}{\fill}
\setlength{\LTright}{\fill}
\begin{longtable}{@{}lrrrrrrrrrr@{}}
 \caption{Full results. Colors show percentile ranks among available languages per column: red indicates lower ranks, white middle ranks, and green higher ranks, using the same scale in each column.\label{tab:full-results}}\\
\toprule
Language & Bak. & HDE & ROE & $\Delta H$ & $F^{(.25)}$ & $F^{(.5)}$ & $F^{(.75)}$ & $F^{(1)}$ & TTR & $|\mathcal{P}|$ \\
\midrule
\endfirsthead
\caption[]{Full results (continued).}\\
\toprule
Language & Bak. & HDE & ROE & $\Delta H$ & $F^{(.25)}$ & $F^{(.5)}$ & $F^{(.75)}$ & $F^{(1)}$ & TTR & $|\mathcal{P}|$ \\
\midrule
\endhead
\midrule
\multicolumn{11}{r}{\emph{Continued on next page}}\\
\endfoot
\bottomrule
\endlastfoot
Abkhaz & \cellcolor{paperGreen!22}$0.50$ & \cellcolor{paperGreen!18}$0.33$ & \cellcolor{paperGreen!37}$0.35$ & \cellcolor{paperGreen!57}$-0.15$ & \cellcolor{paperRed!18}$0.13$ & \cellcolor{paperRed!23}$0.20$ & \cellcolor{paperRed!45}$0.27$ & \cellcolor{paperRed!56}$0.27$ & $0.60$ & 104 \\
Afrikaans & -- & \cellcolor{paperGreen!17}$0.33$ & \cellcolor{paperGreen!20}$0.29$ & \cellcolor{paperRed!57}$-1.55$ & \cellcolor{paperRed!32}$0.12$ & \cellcolor{paperRed!28}$0.19$ & \cellcolor{paperRed!22}$0.31$ & \cellcolor{paperGreen!4}$0.46$ & $0.17$ & 18,243 \\
Akkadian & -- & \cellcolor{paperRed!47}$0.11$ & \cellcolor{paperRed!55}$0.11$ & \cellcolor{paperRed!60}$-3.13$ & \cellcolor{paperRed!59}$0.07$ & \cellcolor{paperRed!57}$0.11$ & \cellcolor{paperRed!57}$0.17$ & \cellcolor{paperRed!55}$0.28$ & $0.23$ & 11,146 \\
Alemannic & -- & \cellcolor{paperGreen!28}$0.36$ & \cellcolor{paperRed!38}$0.15$ & \cellcolor{paperRed!4}$-0.52$ & \cellcolor{paperRed!54}$0.09$ & \cellcolor{paperRed!51}$0.15$ & \cellcolor{paperRed!51}$0.26$ & \cellcolor{paperRed!4}$0.45$ & $0.30$ & 17,676 \\
Amharic & -- & \cellcolor{paperGreen!43}$0.43$ & \cellcolor{paperGreen!38}$0.36$ & \cellcolor{paperRed!40}$-0.88$ & \cellcolor{paperGreen!49}$0.18$ & \cellcolor{paperGreen!41}$0.25$ & \cellcolor{paperRed!11}$0.32$ & \cellcolor{paperRed!45}$0.35$ & $0.16$ & 2,371 \\
Ancient Greek & -- & \cellcolor{paperGreen!60}$0.71$ & \cellcolor{paperGreen!60}$0.82$ & \cellcolor{paperGreen!46}$-0.22$ & \cellcolor{paperRed!4}$0.14$ & \cellcolor{paperRed!3}$0.22$ & $0.33$ & \cellcolor{paperRed!26}$0.42$ & $0.47$ & 1,794 \\
Ancient Hebrew & -- & \cellcolor{paperRed!36}$0.16$ & \cellcolor{paperRed!43}$0.14$ & \cellcolor{paperRed!37}$-0.83$ & \cellcolor{paperRed!27}$0.12$ & \cellcolor{paperRed!24}$0.20$ & \cellcolor{paperRed!19}$0.32$ & \cellcolor{paperRed!31}$0.41$ & $0.32$ & 14,292 \\
Arabic & -- & \cellcolor{paperRed!28}$0.18$ & \cellcolor{paperRed!45}$0.14$ & \cellcolor{paperGreen!31}$-0.34$ & \cellcolor{paperRed!1}$0.14$ & \cellcolor{paperGreen!9}$0.22$ & \cellcolor{paperGreen!31}$0.36$ & \cellcolor{paperGreen!29}$0.49$ & $0.37$ & 8,269 \\
Bambara & -- & \cellcolor{paperRed!31}$0.18$ & \cellcolor{paperRed!33}$0.16$ & \cellcolor{paperRed!51}$-1.25$ & \cellcolor{paperGreen!47}$0.17$ & \cellcolor{paperGreen!45}$0.25$ & \cellcolor{paperGreen!33}$0.36$ & \cellcolor{paperGreen!6}$0.46$ & $0.08$ & 6,164 \\
Basque & \cellcolor{paperGreen!41}$0.60$ & \cellcolor{paperGreen!56}$0.53$ & \cellcolor{paperGreen!59}$0.56$ & \cellcolor{paperGreen!54}$-0.20$ & \cellcolor{paperGreen!57}$0.19$ & \cellcolor{paperGreen!59}$0.29$ & \cellcolor{paperGreen!59}$0.41$ & \cellcolor{paperGreen!24}$0.48$ & $0.49$ & 860 \\
Belarusian & -- & \cellcolor{paperGreen!34}$0.38$ & \cellcolor{paperGreen!36}$0.35$ & \cellcolor{paperGreen!37}$-0.29$ & \cellcolor{paperRed!11}$0.13$ & \cellcolor{paperRed!13}$0.21$ & \cellcolor{paperRed!9}$0.32$ & \cellcolor{paperGreen!1}$0.45$ & $0.45$ & 4,333 \\
Bororo & -- & \cellcolor{paperGreen!9}$0.29$ & \cellcolor{paperGreen!28}$0.33$ & \cellcolor{paperGreen!27}$-0.37$ & \cellcolor{paperGreen!22}$0.15$ & \cellcolor{paperGreen!1}$0.22$ & \cellcolor{paperRed!34}$0.30$ & \cellcolor{paperRed!46}$0.35$ & $0.33$ & 1,381 \\
Bulgarian & \cellcolor{paperGreen!41}$0.60$ & \cellcolor{paperGreen!36}$0.39$ & \cellcolor{paperGreen!23}$0.31$ & \cellcolor{paperRed!23}$-0.70$ & \cellcolor{paperGreen!52}$0.18$ & \cellcolor{paperGreen!36}$0.25$ & \cellcolor{paperRed!6}$0.33$ & \cellcolor{paperRed!40}$0.37$ & $0.37$ & 1,152 \\
Cantonese & -- & \cellcolor{paperRed!22}$0.19$ & \cellcolor{paperRed!11}$0.20$ & \cellcolor{paperRed!28}$-0.74$ & \cellcolor{paperGreen!24}$0.15$ & \cellcolor{paperGreen!20}$0.23$ & \cellcolor{paperGreen!14}$0.35$ & \cellcolor{paperRed!8}$0.45$ & $0.14$ & 7,543 \\
Catalan & \cellcolor{paperRed!2}$0.40$ & \cellcolor{paperRed!29}$0.18$ & \cellcolor{paperRed!18}$0.20$ & \cellcolor{paperRed!47}$-1.11$ & \cellcolor{paperRed!42}$0.11$ & \cellcolor{paperRed!42}$0.18$ & \cellcolor{paperRed!28}$0.31$ & \cellcolor{paperGreen!51}$0.50$ & $0.18$ & 35,719 \\
Chinese Simplified & -- & \cellcolor{paperRed!40}$0.14$ & \cellcolor{paperRed!24}$0.19$ & \cellcolor{paperGreen!1}$-0.49$ & \cellcolor{paperRed!37}$0.11$ & \cellcolor{paperRed!37}$0.19$ & \cellcolor{paperRed!26}$0.31$ & \cellcolor{paperRed!1}$0.45$ & $0.33$ & 13,776 \\
Chinese Traditional & -- & \cellcolor{paperRed!38}$0.14$ & \cellcolor{paperRed!22}$0.19$ & \cellcolor{paperGreen!3}$-0.49$ & \cellcolor{paperRed!40}$0.11$ & \cellcolor{paperRed!38}$0.19$ & \cellcolor{paperRed!27}$0.31$ & \cellcolor{paperRed!3}$0.45$ & $0.33$ & 13,754 \\
Chintang & -- & \cellcolor{paperRed!10}$0.23$ & $0.25$ & \cellcolor{paperGreen!28}$-0.36$ & \cellcolor{paperGreen!56}$0.19$ & \cellcolor{paperGreen!46}$0.25$ & \cellcolor{paperRed!24}$0.31$ & \cellcolor{paperRed!49}$0.32$ & $0.46$ & 348 \\
Chukchi & -- & \cellcolor{paperGreen!42}$0.43$ & \cellcolor{paperGreen!40}$0.36$ & \cellcolor{paperGreen!51}$-0.20$ & \cellcolor{paperGreen!60}$0.27$ & \cellcolor{paperGreen!60}$0.35$ & \cellcolor{paperGreen!60}$0.43$ & \cellcolor{paperRed!9}$0.45$ & $0.44$ & 498 \\
Coptic & -- & \cellcolor{paperRed!34}$0.17$ & \cellcolor{paperRed!51}$0.12$ & \cellcolor{paperRed!45}$-1.01$ & \cellcolor{paperGreen!3}$0.14$ & \cellcolor{paperGreen!8}$0.22$ & \cellcolor{paperGreen!27}$0.36$ & \cellcolor{paperGreen!60}$0.53$ & $0.08$ & 20,506 \\
Croatian & -- & \cellcolor{paperGreen!3}$0.27$ & \cellcolor{paperRed!32}$0.17$ & \cellcolor{paperRed!36}$-0.81$ & \cellcolor{paperRed!15}$0.13$ & \cellcolor{paperRed!17}$0.20$ & \cellcolor{paperRed!8}$0.32$ & \cellcolor{paperGreen!14}$0.47$ & $0.35$ & 8,715 \\
Czech & -- & \cellcolor{paperGreen!32}$0.37$ & \cellcolor{paperGreen!27}$0.33$ & \cellcolor{paperRed!27}$-0.74$ & \cellcolor{paperRed!31}$0.12$ & \cellcolor{paperRed!29}$0.19$ & \cellcolor{paperRed!31}$0.31$ & \cellcolor{paperRed!17}$0.43$ & $0.38$ & 4,028 \\
Danish & \cellcolor{paperRed!32}$0.30$ & \cellcolor{paperGreen!23}$0.35$ & \cellcolor{paperGreen!18}$0.29$ & \cellcolor{paperGreen!45}$-0.23$ & \cellcolor{paperGreen!18}$0.15$ & \cellcolor{paperGreen!14}$0.23$ & \cellcolor{paperGreen!34}$0.36$ & \cellcolor{paperGreen!57}$0.52$ & $0.34$ & 6,514 \\
Dutch & \cellcolor{paperRed!2}$0.40$ & \cellcolor{paperRed!9}$0.23$ & \cellcolor{paperRed!56}$0.11$ & \cellcolor{paperRed!34}$-0.81$ & \cellcolor{paperGreen!6}$0.14$ & \cellcolor{paperGreen!15}$0.23$ & \cellcolor{paperGreen!36}$0.36$ & \cellcolor{paperGreen!8}$0.46$ & $0.36$ & 3,675 \\
Eastern Armenian & -- & \cellcolor{paperGreen!33}$0.37$ & \cellcolor{paperGreen!31}$0.34$ & \cellcolor{paperRed!46}$-1.08$ & \cellcolor{paperRed!50}$0.09$ & \cellcolor{paperRed!49}$0.16$ & \cellcolor{paperRed!49}$0.26$ & \cellcolor{paperRed!24}$0.42$ & $0.39$ & 6,893 \\
Egyptian & -- & \cellcolor{paperRed!46}$0.12$ & \cellcolor{paperRed!59}$0.10$ & \cellcolor{paperRed!56}$-1.40$ & \cellcolor{paperGreen!37}$0.16$ & \cellcolor{paperGreen!31}$0.24$ & \cellcolor{paperGreen!5}$0.34$ & \cellcolor{paperRed!32}$0.39$ & $0.15$ & 5,781 \\
English & \cellcolor{paperRed!2}$0.40$ & \cellcolor{paperRed!37}$0.15$ & \cellcolor{paperRed!54}$0.11$ & \cellcolor{paperRed!50}$-1.20$ & \cellcolor{paperRed!24}$0.12$ & \cellcolor{paperRed!15}$0.20$ & \cellcolor{paperGreen!8}$0.34$ & \cellcolor{paperGreen!55}$0.51$ & $0.22$ & 15,306 \\
Erzya & -- & \cellcolor{paperGreen!57}$0.53$ & \cellcolor{paperGreen!57}$0.53$ & \cellcolor{paperGreen!59}$-0.13$ & \cellcolor{paperGreen!50}$0.18$ & \cellcolor{paperGreen!57}$0.28$ & \cellcolor{paperGreen!55}$0.39$ & \cellcolor{paperRed!18}$0.43$ & $0.51$ & 253 \\
Estonian & \cellcolor{paperRed!2}$0.40$ & \cellcolor{paperGreen!55}$0.52$ & \cellcolor{paperGreen!50}$0.44$ & \cellcolor{paperGreen!49}$-0.21$ & \cellcolor{paperGreen!20}$0.15$ & \cellcolor{paperGreen!24}$0.24$ & \cellcolor{paperGreen!43}$0.37$ & \cellcolor{paperGreen!33}$0.49$ & $0.40$ & 3,014 \\
Faroese & \cellcolor{paperRed!2}$0.40$ & \cellcolor{paperGreen!6}$0.28$ & \cellcolor{paperRed!34}$0.16$ & \cellcolor{paperGreen!8}$-0.47$ & \cellcolor{paperGreen!59}$0.19$ & \cellcolor{paperGreen!56}$0.27$ & \cellcolor{paperGreen!22}$0.35$ & \cellcolor{paperRed!34}$0.39$ & $0.39$ & 912 \\
Finnish & \cellcolor{paperGreen!41}$0.60$ & \cellcolor{paperGreen!47}$0.47$ & \cellcolor{paperGreen!56}$0.52$ & \cellcolor{paperGreen!18}$-0.43$ & \cellcolor{paperGreen!46}$0.17$ & \cellcolor{paperGreen!40}$0.25$ & \cellcolor{paperGreen!15}$0.35$ & \cellcolor{paperRed!42}$0.36$ & $0.71$ & 178 \\
French & \cellcolor{paperRed!60}$0.10$ & \cellcolor{paperRed!32}$0.18$ & \cellcolor{paperGreen!3}$0.25$ & \cellcolor{paperRed!54}$-1.35$ & \cellcolor{paperRed!46}$0.10$ & \cellcolor{paperRed!46}$0.17$ & \cellcolor{paperRed!41}$0.29$ & \cellcolor{paperGreen!36}$0.49$ & $0.16$ & 27,662 \\
Galician & \cellcolor{paperRed!52}$0.20$ & \cellcolor{paperRed!14}$0.22$ & \cellcolor{paperRed!3}$0.24$ & \cellcolor{paperGreen!24}$-0.38$ & \cellcolor{paperGreen!11}$0.14$ & \cellcolor{paperGreen!17}$0.23$ & \cellcolor{paperGreen!47}$0.37$ & \cellcolor{paperGreen!38}$0.49$ & $0.28$ & 13,436 \\
Georgian & \cellcolor{paperGreen!22}$0.50$ & \cellcolor{paperGreen!1}$0.26$ & \cellcolor{paperGreen!17}$0.28$ & \cellcolor{paperRed!18}$-0.68$ & \cellcolor{paperRed!6}$0.14$ & \cellcolor{paperRed!6}$0.21$ & \cellcolor{paperGreen!4}$0.33$ & \cellcolor{paperRed!14}$0.44$ & $0.41$ & 5,134 \\
German & \cellcolor{paperRed!2}$0.40$ & \cellcolor{paperGreen!19}$0.33$ & \cellcolor{paperRed!10}$0.21$ & \cellcolor{paperRed!41}$-0.89$ & \cellcolor{paperRed!5}$0.14$ & \cellcolor{paperRed!1}$0.22$ & \cellcolor{paperGreen!10}$0.34$ & \cellcolor{paperGreen!10}$0.47$ & $0.27$ & 8,224 \\
Gothic & \cellcolor{paperGreen!54}$0.70$ & \cellcolor{paperGreen!52}$0.50$ & \cellcolor{paperGreen!52}$0.47$ & \cellcolor{paperRed!11}$-0.61$ & \cellcolor{paperGreen!40}$0.17$ & \cellcolor{paperGreen!37}$0.25$ & \cellcolor{paperGreen!28}$0.36$ & \cellcolor{paperRed!28}$0.41$ & $0.27$ & 3,013 \\
Guajajara & -- & \cellcolor{paperGreen!51}$0.49$ & \cellcolor{paperGreen!55}$0.51$ & \cellcolor{paperRed!29}$-0.74$ & \cellcolor{paperGreen!17}$0.15$ & \cellcolor{paperRed!9}$0.21$ & \cellcolor{paperRed!46}$0.27$ & \cellcolor{paperRed!54}$0.30$ & $0.18$ & 1,609 \\
Haitian Creole & -- & \cellcolor{paperGreen!26}$0.35$ & \cellcolor{paperGreen!29}$0.34$ & \cellcolor{paperRed!55}$-1.38$ & \cellcolor{paperGreen!14}$0.15$ & \cellcolor{paperGreen!22}$0.23$ & \cellcolor{paperGreen!41}$0.37$ & \cellcolor{paperGreen!56}$0.51$ & $0.10$ & 15,025 \\
Hausa & -- & \cellcolor{paperRed!20}$0.19$ & \cellcolor{paperRed!37}$0.15$ & \cellcolor{paperRed!42}$-0.95$ & \cellcolor{paperGreen!15}$0.15$ & \cellcolor{paperGreen!11}$0.23$ & \cellcolor{paperGreen!6}$0.34$ & \cellcolor{paperRed!20}$0.42$ & $0.15$ & 9,141 \\
Hebrew & -- & \cellcolor{paperRed!23}$0.19$ & \cellcolor{paperRed!36}$0.15$ & \cellcolor{paperRed!13}$-0.61$ & \cellcolor{paperRed!14}$0.13$ & \cellcolor{paperRed!14}$0.20$ & \cellcolor{paperRed!5}$0.33$ & \cellcolor{paperGreen!20}$0.47$ & $0.19$ & 18,153 \\
Highland Puebla Nahuatl & -- & \cellcolor{paperGreen!31}$0.37$ & \cellcolor{paperGreen!43}$0.39$ & \cellcolor{paperGreen!11}$-0.47$ & \cellcolor{paperGreen!55}$0.18$ & \cellcolor{paperGreen!55}$0.27$ & \cellcolor{paperGreen!46}$0.37$ & \cellcolor{paperRed!6}$0.45$ & $0.25$ & 2,259 \\
Hindi & -- & \cellcolor{paperRed!52}$0.08$ & \cellcolor{paperRed!50}$0.12$ & \cellcolor{paperRed!31}$-0.75$ & \cellcolor{paperRed!23}$0.12$ & \cellcolor{paperRed!26}$0.20$ & \cellcolor{paperRed!32}$0.31$ & \cellcolor{paperRed!27}$0.41$ & $0.24$ & 8,483 \\
Hungarian & -- & \cellcolor{paperGreen!20}$0.34$ & \cellcolor{paperGreen!9}$0.27$ & \cellcolor{paperGreen!26}$-0.38$ & \cellcolor{paperRed!13}$0.13$ & \cellcolor{paperRed!8}$0.21$ & \cellcolor{paperGreen!19}$0.35$ & \cellcolor{paperGreen!31}$0.49$ & $0.43$ & 8,951 \\
Icelandic & \cellcolor{paperRed!2}$0.40$ & \cellcolor{paperGreen!27}$0.36$ & \cellcolor{paperGreen!6}$0.26$ & \cellcolor{paperGreen!41}$-0.26$ & \cellcolor{paperGreen!5}$0.14$ & \cellcolor{paperGreen!5}$0.22$ & \cellcolor{paperGreen!24}$0.35$ & \cellcolor{paperGreen!47}$0.50$ & $0.35$ & 9,317 \\
Indonesian & -- & \cellcolor{paperGreen!4}$0.28$ & \cellcolor{paperGreen!19}$0.29$ & \cellcolor{paperRed!49}$-1.17$ & \cellcolor{paperRed!45}$0.10$ & \cellcolor{paperRed!43}$0.18$ & \cellcolor{paperRed!36}$0.30$ & \cellcolor{paperRed!10}$0.44$ & $0.18$ & 22,487 \\
Irish & \cellcolor{paperRed!52}$0.20$ & \cellcolor{paperRed!26}$0.19$ & \cellcolor{paperRed!46}$0.14$ & \cellcolor{paperGreen!9}$-0.47$ & \cellcolor{paperRed!57}$0.08$ & \cellcolor{paperRed!55}$0.14$ & \cellcolor{paperRed!52}$0.24$ & \cellcolor{paperGreen!9}$0.46$ & $0.27$ & 25,630 \\
Italian & \cellcolor{paperRed!32}$0.30$ & \cellcolor{paperRed!11}$0.22$ & \cellcolor{paperGreen!24}$0.31$ & \cellcolor{paperRed!52}$-1.27$ & \cellcolor{paperRed!43}$0.10$ & \cellcolor{paperRed!45}$0.18$ & \cellcolor{paperRed!38}$0.30$ & \cellcolor{paperGreen!49}$0.50$ & $0.19$ & 27,218 \\
Japanese & -- & \cellcolor{paperRed!60}$0.03$ & \cellcolor{paperRed!42}$0.14$ & \cellcolor{paperGreen!5}$-0.48$ & \cellcolor{paperRed!33}$0.12$ & \cellcolor{paperRed!32}$0.19$ & \cellcolor{paperRed!17}$0.32$ & \cellcolor{paperGreen!18}$0.47$ & $0.26$ & 18,355 \\
Javanese & -- & \cellcolor{paperGreen!8}$0.29$ & \cellcolor{paperGreen!11}$0.27$ & \cellcolor{paperGreen!4}$-0.49$ & \cellcolor{paperGreen!32}$0.16$ & \cellcolor{paperGreen!38}$0.25$ & \cellcolor{paperGreen!51}$0.38$ & \cellcolor{paperGreen!17}$0.47$ & $0.31$ & 4,168 \\
Kazakh & -- & \cellcolor{paperRed!56}$0.06$ & \cellcolor{paperRed!23}$0.19$ & \cellcolor{paperGreen!47}$-0.21$ & \cellcolor{paperRed!49}$0.10$ & \cellcolor{paperRed!52}$0.15$ & \cellcolor{paperRed!55}$0.23$ & \cellcolor{paperRed!51}$0.31$ & $0.53$ & 454 \\
Khoekhoe & -- & \cellcolor{paperGreen!49}$0.47$ & \cellcolor{paperGreen!41}$0.37$ & \cellcolor{paperRed!9}$-0.56$ & \cellcolor{paperGreen!45}$0.17$ & \cellcolor{paperGreen!50}$0.26$ & \cellcolor{paperGreen!42}$0.37$ & \cellcolor{paperRed!11}$0.44$ & $0.19$ & 2,392 \\
K'iche' & -- & \cellcolor{paperGreen!13}$0.31$ & \cellcolor{paperGreen!4}$0.25$ & \cellcolor{paperGreen!10}$-0.47$ & \cellcolor{paperGreen!51}$0.18$ & \cellcolor{paperGreen!47}$0.26$ & \cellcolor{paperRed!4}$0.33$ & \cellcolor{paperRed!37}$0.38$ & $0.30$ & 876 \\
Korean & -- & \cellcolor{paperRed!57}$0.05$ & \cellcolor{paperRed!5}$0.23$ & \cellcolor{paperGreen!33}$-0.32$ & \cellcolor{paperRed!51}$0.09$ & \cellcolor{paperRed!56}$0.13$ & \cellcolor{paperRed!56}$0.19$ & \cellcolor{paperRed!57}$0.23$ & $0.52$ & 428 \\
Kyrgyz & -- & \cellcolor{paperRed!59}$0.05$ & \cellcolor{paperRed!40}$0.15$ & \cellcolor{paperRed!38}$-0.85$ & \cellcolor{paperRed!60}$0.03$ & \cellcolor{paperRed!60}$0.05$ & \cellcolor{paperRed!60}$0.06$ & \cellcolor{paperRed!60}$0.06$ & $0.41$ & 854 \\
Latin & \cellcolor{paperGreen!60}$0.90$ & \cellcolor{paperGreen!54}$0.51$ & \cellcolor{paperGreen!45}$0.41$ & \cellcolor{paperRed!19}$-0.69$ & \cellcolor{paperGreen!31}$0.16$ & \cellcolor{paperGreen!29}$0.24$ & \cellcolor{paperGreen!29}$0.36$ & \cellcolor{paperGreen!13}$0.47$ & $0.17$ & 7,921 \\
Latvian & \cellcolor{paperGreen!22}$0.50$ & \cellcolor{paperGreen!37}$0.41$ & \cellcolor{paperGreen!46}$0.41$ & \cellcolor{paperGreen!56}$-0.16$ & \cellcolor{paperGreen!29}$0.16$ & \cellcolor{paperGreen!34}$0.25$ & \cellcolor{paperGreen!54}$0.39$ & \cellcolor{paperGreen!42}$0.50$ & $0.48$ & 2,206 \\
Lithuanian & \cellcolor{paperRed!32}$0.30$ & \cellcolor{paperGreen!40}$0.42$ & \cellcolor{paperGreen!49}$0.43$ & \cellcolor{paperGreen!29}$-0.34$ & \cellcolor{paperRed!36}$0.11$ & \cellcolor{paperRed!40}$0.19$ & \cellcolor{paperRed!33}$0.30$ & \cellcolor{paperGreen!3}$0.45$ & $0.44$ & 5,881 \\
Maltese & \cellcolor{paperRed!32}$0.30$ & \cellcolor{paperRed!3}$0.25$ & \cellcolor{paperRed!28}$0.18$ & \cellcolor{paperRed!20}$-0.70$ & \cellcolor{paperRed!41}$0.11$ & \cellcolor{paperRed!41}$0.18$ & \cellcolor{paperRed!29}$0.31$ & \cellcolor{paperGreen!34}$0.49$ & $0.25$ & 14,808 \\
Manx & \cellcolor{paperRed!32}$0.30$ & \cellcolor{paperRed!49}$0.10$ & \cellcolor{paperRed!60}$0.09$ & \cellcolor{paperRed!3}$-0.50$ & \cellcolor{paperGreen!8}$0.14$ & $0.22$ & \cellcolor{paperRed!1}$0.33$ & $0.45$ & $0.22$ & 3,278 \\
Modern Greek & \cellcolor{paperGreen!41}$0.60$ & \cellcolor{paperRed!5}$0.24$ & \cellcolor{paperRed!41}$0.14$ & \cellcolor{paperGreen!17}$-0.43$ & \cellcolor{paperGreen!34}$0.16$ & \cellcolor{paperGreen!43}$0.25$ & \cellcolor{paperGreen!52}$0.38$ & \cellcolor{paperGreen!37}$0.49$ & $0.27$ & 4,076 \\
Naija & -- & \cellcolor{paperRed!45}$0.12$ & \cellcolor{paperRed!57}$0.10$ & \cellcolor{paperRed!33}$-0.78$ & \cellcolor{paperGreen!28}$0.16$ & \cellcolor{paperGreen!26}$0.24$ & \cellcolor{paperGreen!11}$0.35$ & \cellcolor{paperRed!19}$0.42$ & $0.09$ & 4,754 \\
Nheengatu & -- & \cellcolor{paperGreen!50}$0.48$ & \cellcolor{paperGreen!47}$0.41$ & \cellcolor{paperGreen!13}$-0.47$ & \cellcolor{paperGreen!41}$0.17$ & \cellcolor{paperGreen!51}$0.26$ & \cellcolor{paperGreen!50}$0.38$ & \cellcolor{paperRed!5}$0.45$ & $0.18$ & 4,190 \\
North Sami & -- & \cellcolor{paperGreen!46}$0.47$ & \cellcolor{paperGreen!54}$0.49$ & \cellcolor{paperGreen!38}$-0.27$ & \cellcolor{paperGreen!54}$0.18$ & \cellcolor{paperGreen!52}$0.26$ & \cellcolor{paperGreen!9}$0.34$ & \cellcolor{paperRed!43}$0.36$ & $0.46$ & 404 \\
Norwegian Bokm{\aa}l & -- & \cellcolor{paperGreen!11}$0.31$ & \cellcolor{paperRed!8}$0.23$ & \cellcolor{paperRed!1}$-0.50$ & \cellcolor{paperGreen!27}$0.15$ & \cellcolor{paperGreen!32}$0.24$ & \cellcolor{paperGreen!45}$0.37$ & \cellcolor{paperGreen!46}$0.50$ & $0.29$ & 5,838 \\
Norwegian Nynorsk & -- & \cellcolor{paperGreen!14}$0.31$ & \cellcolor{paperGreen!1}$0.25$ & \cellcolor{paperGreen!23}$-0.39$ & \cellcolor{paperGreen!26}$0.15$ & \cellcolor{paperGreen!28}$0.24$ & \cellcolor{paperGreen!49}$0.37$ & \cellcolor{paperGreen!52}$0.50$ & $0.26$ & 7,840 \\
Occitan & -- & \cellcolor{paperRed!8}$0.23$ & \cellcolor{paperRed!9}$0.22$ & \cellcolor{paperGreen!32}$-0.32$ & \cellcolor{paperRed!8}$0.13$ & \cellcolor{paperRed!4}$0.22$ & \cellcolor{paperGreen!13}$0.35$ & \cellcolor{paperGreen!26}$0.48$ & $0.29$ & 9,214 \\
Ottoman Turkish & -- & \cellcolor{paperRed!17}$0.21$ & \cellcolor{paperGreen!33}$0.34$ & \cellcolor{paperGreen!55}$-0.19$ & \cellcolor{paperRed!28}$0.12$ & \cellcolor{paperRed!33}$0.19$ & \cellcolor{paperRed!40}$0.29$ & \cellcolor{paperRed!33}$0.39$ & $0.57$ & 2,374 \\
Persian & -- & \cellcolor{paperRed!24}$0.19$ & \cellcolor{paperRed!15}$0.20$ & \cellcolor{paperGreen!6}$-0.48$ & \cellcolor{paperRed!3}$0.14$ & \cellcolor{paperGreen!3}$0.22$ & \cellcolor{paperGreen!20}$0.35$ & \cellcolor{paperGreen!45}$0.50$ & $0.20$ & 11,641 \\
Polish & \cellcolor{paperGreen!41}$0.60$ & \cellcolor{paperRed!18}$0.21$ & \cellcolor{paperGreen!15}$0.27$ & \cellcolor{paperRed!43}$-0.96$ & \cellcolor{paperRed!20}$0.13$ & \cellcolor{paperRed!31}$0.19$ & \cellcolor{paperRed!50}$0.26$ & \cellcolor{paperRed!52}$0.30$ & $0.20$ & 3,571 \\
Pomak & -- & \cellcolor{paperGreen!45}$0.46$ & \cellcolor{paperGreen!13}$0.27$ & $-0.50$ & \cellcolor{paperGreen!23}$0.15$ & \cellcolor{paperGreen!23}$0.24$ & \cellcolor{paperGreen!37}$0.36$ & \cellcolor{paperGreen!28}$0.48$ & $0.27$ & 8,159 \\
Portuguese & \cellcolor{paperRed!32}$0.30$ & \cellcolor{paperRed!15}$0.21$ & \cellcolor{paperRed!13}$0.20$ & \cellcolor{paperGreen!40}$-0.27$ & \cellcolor{paperGreen!36}$0.16$ & \cellcolor{paperGreen!49}$0.26$ & \cellcolor{paperGreen!56}$0.40$ & \cellcolor{paperGreen!50}$0.50$ & $0.30$ & 8,533 \\
Romanian & -- & \cellcolor{paperGreen!22}$0.34$ & \cellcolor{paperGreen!22}$0.29$ & \cellcolor{paperRed!8}$-0.55$ & \cellcolor{paperRed!22}$0.12$ & \cellcolor{paperRed!20}$0.20$ & \cellcolor{paperRed!13}$0.32$ & \cellcolor{paperGreen!32}$0.49$ & $0.27$ & 12,841 \\
Russian & \cellcolor{paperGreen!54}$0.70$ & \cellcolor{paperRed!4}$0.24$ & \cellcolor{paperRed!14}$0.20$ & \cellcolor{paperGreen!34}$-0.30$ & $0.14$ & \cellcolor{paperGreen!13}$0.23$ & \cellcolor{paperGreen!40}$0.37$ & \cellcolor{paperGreen!27}$0.48$ & $0.47$ & 4,008 \\
Sanskrit Vedic & -- & \cellcolor{paperGreen!59}$0.54$ & \cellcolor{paperGreen!34}$0.34$ & \cellcolor{paperGreen!20}$-0.40$ & \cellcolor{paperGreen!42}$0.17$ & \cellcolor{paperGreen!33}$0.25$ & \cellcolor{paperGreen!17}$0.35$ & \cellcolor{paperRed!36}$0.39$ & $0.49$ & 1,910 \\
Scottish Gaelic & \cellcolor{paperRed!32}$0.30$ & \cellcolor{paperRed!41}$0.14$ & \cellcolor{paperRed!52}$0.11$ & \cellcolor{paperRed!26}$-0.74$ & \cellcolor{paperGreen!19}$0.15$ & \cellcolor{paperGreen!19}$0.23$ & \cellcolor{paperGreen!23}$0.35$ & \cellcolor{paperGreen!23}$0.48$ & $0.13$ & 5,245 \\
Serbian & -- & $0.26$ & \cellcolor{paperRed!31}$0.17$ & \cellcolor{paperRed!32}$-0.76$ & \cellcolor{paperRed!10}$0.13$ & \cellcolor{paperRed!10}$0.21$ & \cellcolor{paperGreen!1}$0.33$ & \cellcolor{paperGreen!15}$0.47$ & $0.34$ & 9,806 \\
Sindhi & -- & \cellcolor{paperRed!19}$0.20$ & \cellcolor{paperRed!19}$0.19$ & \cellcolor{paperRed!17}$-0.67$ & \cellcolor{paperRed!9}$0.13$ & \cellcolor{paperRed!5}$0.21$ & \cellcolor{paperRed!3}$0.33$ & \cellcolor{paperRed!22}$0.42$ & $0.21$ & 6,831 \\
Slovak & \cellcolor{paperGreen!22}$0.50$ & \cellcolor{paperGreen!41}$0.42$ & \cellcolor{paperGreen!32}$0.34$ & \cellcolor{paperGreen!14}$-0.46$ & \cellcolor{paperGreen!43}$0.17$ & \cellcolor{paperGreen!42}$0.25$ & \cellcolor{paperGreen!32}$0.36$ & \cellcolor{paperRed!15}$0.44$ & $0.54$ & 847 \\
Slovenian & -- & \cellcolor{paperGreen!15}$0.32$ & \cellcolor{paperRed!1}$0.24$ & \cellcolor{paperGreen!36}$-0.30$ & \cellcolor{paperRed!29}$0.12$ & \cellcolor{paperRed!18}$0.20$ & \cellcolor{paperGreen!3}$0.33$ & \cellcolor{paperGreen!41}$0.50$ & $0.43$ & 7,344 \\
Spanish & \cellcolor{paperRed!32}$0.30$ & \cellcolor{paperRed!6}$0.24$ & \cellcolor{paperGreen!5}$0.25$ & \cellcolor{paperGreen!22}$-0.40$ & \cellcolor{paperRed!34}$0.12$ & \cellcolor{paperRed!34}$0.19$ & \cellcolor{paperRed!20}$0.32$ & \cellcolor{paperGreen!22}$0.47$ & $0.30$ & 22,955 \\
Swedish & \cellcolor{paperRed!2}$0.40$ & \cellcolor{paperGreen!10}$0.29$ & \cellcolor{paperRed!6}$0.23$ & \cellcolor{paperRed!24}$-0.73$ & \cellcolor{paperGreen!10}$0.14$ & \cellcolor{paperGreen!10}$0.23$ & \cellcolor{paperGreen!18}$0.35$ & \cellcolor{paperGreen!19}$0.47$ & $0.24$ & 8,259 \\
Tamil & -- & \cellcolor{paperRed!55}$0.07$ & \cellcolor{paperGreen!8}$0.26$ & \cellcolor{paperRed!10}$-0.60$ & \cellcolor{paperRed!56}$0.09$ & \cellcolor{paperRed!54}$0.15$ & \cellcolor{paperRed!54}$0.24$ & \cellcolor{paperRed!50}$0.31$ & $0.41$ & 2,306 \\
Telugu & -- & \cellcolor{paperRed!51}$0.09$ & \cellcolor{paperGreen!14}$0.27$ & \cellcolor{paperGreen!43}$-0.24$ & \cellcolor{paperRed!55}$0.09$ & \cellcolor{paperRed!59}$0.11$ & \cellcolor{paperRed!59}$0.11$ & \cellcolor{paperRed!59}$0.11$ & $0.45$ & 129 \\
Thai & -- & \cellcolor{paperRed!13}$0.22$ & \cellcolor{paperGreen!10}$0.27$ & \cellcolor{paperRed!22}$-0.70$ & \cellcolor{paperRed!47}$0.10$ & \cellcolor{paperRed!47}$0.17$ & \cellcolor{paperRed!42}$0.29$ & \cellcolor{paperGreen!59}$0.52$ & $0.16$ & 26,287 \\
Turkish & \cellcolor{paperRed!52}$0.20$ & \cellcolor{paperRed!54}$0.08$ & \cellcolor{paperRed!29}$0.17$ & \cellcolor{paperGreen!60}$-0.11$ & \cellcolor{paperRed!19}$0.13$ & \cellcolor{paperRed!19}$0.20$ & \cellcolor{paperRed!18}$0.32$ & \cellcolor{paperRed!23}$0.42$ & $0.58$ & 1,279 \\
Ukrainian & -- & \cellcolor{paperGreen!24}$0.35$ & \cellcolor{paperGreen!26}$0.31$ & \cellcolor{paperGreen!50}$-0.21$ & \cellcolor{paperRed!26}$0.12$ & \cellcolor{paperRed!22}$0.20$ & \cellcolor{paperRed!14}$0.32$ & \cellcolor{paperGreen!11}$0.47$ & $0.46$ & 4,416 \\
Urdu & -- & \cellcolor{paperRed!43}$0.13$ & \cellcolor{paperRed!26}$0.18$ & \cellcolor{paperRed!59}$-2.40$ & \cellcolor{paperRed!52}$0.09$ & \cellcolor{paperRed!50}$0.15$ & \cellcolor{paperRed!47}$0.27$ & \cellcolor{paperRed!13}$0.44$ & $0.17$ & 27,581 \\
Uyghur & -- & \cellcolor{paperRed!50}$0.09$ & \cellcolor{paperRed!20}$0.19$ & \cellcolor{paperGreen!42}$-0.25$ & \cellcolor{paperRed!17}$0.13$ & \cellcolor{paperRed!27}$0.20$ & \cellcolor{paperRed!43}$0.29$ & \cellcolor{paperRed!47}$0.34$ & $0.51$ & 875 \\
Uzbek & -- & \cellcolor{paperRed!42}$0.14$ & \cellcolor{paperRed!4}$0.23$ & \cellcolor{paperGreen!52}$-0.20$ & \cellcolor{paperGreen!4}$0.14$ & \cellcolor{paperRed!11}$0.21$ & \cellcolor{paperRed!37}$0.30$ & \cellcolor{paperRed!41}$0.37$ & $0.56$ & 498 \\
Vietnamese & -- & \cellcolor{paperRed!1}$0.25$ & \cellcolor{paperRed!17}$0.20$ & \cellcolor{paperGreen!15}$-0.45$ & \cellcolor{paperGreen!38}$0.16$ & \cellcolor{paperGreen!54}$0.26$ & \cellcolor{paperGreen!57}$0.40$ & \cellcolor{paperGreen!43}$0.50$ & $0.22$ & 5,122 \\
Welsh & -- & \cellcolor{paperRed!27}$0.18$ & \cellcolor{paperRed!27}$0.18$ & \cellcolor{paperRed!6}$-0.55$ & \cellcolor{paperGreen!1}$0.14$ & \cellcolor{paperGreen!6}$0.22$ & \cellcolor{paperGreen!26}$0.36$ & \cellcolor{paperGreen!54}$0.50$ & $0.23$ & 11,436 \\
Western Armenian & -- & \cellcolor{paperGreen!38}$0.41$ & \cellcolor{paperGreen!51}$0.47$ & \cellcolor{paperGreen!19}$-0.41$ & \cellcolor{paperRed!38}$0.11$ & \cellcolor{paperRed!36}$0.19$ & \cellcolor{paperRed!23}$0.31$ & \cellcolor{paperGreen!5}$0.46$ & $0.39$ & 7,417 \\
Wolof & -- & \cellcolor{paperGreen!29}$0.36$ & \cellcolor{paperGreen!42}$0.38$ & \cellcolor{paperRed!15}$-0.65$ & \cellcolor{paperGreen!13}$0.14$ & \cellcolor{paperGreen!18}$0.23$ & \cellcolor{paperGreen!38}$0.36$ & \cellcolor{paperGreen!40}$0.50$ & $0.18$ & 10,899 \\
Yiddish & \cellcolor{paperGreen!22}$0.50$ & \cellcolor{paperRed!33}$0.18$ & \cellcolor{paperRed!49}$0.13$ & \cellcolor{paperRed!5}$-0.55$ & \cellcolor{paperGreen!33}$0.16$ & \cellcolor{paperGreen!27}$0.24$ & \cellcolor{paperRed!15}$0.32$ & \cellcolor{paperRed!38}$0.38$ & $0.29$ & 1,084 \\
Zaar & -- & \cellcolor{paperGreen!5}$0.28$ & \cellcolor{paperRed!47}$0.13$ & \cellcolor{paperRed!14}$-0.63$ & \cellcolor{paperGreen!9}$0.14$ & \cellcolor{paperGreen!4}$0.22$ & \cellcolor{paperRed!10}$0.32$ & \cellcolor{paperRed!29}$0.41$ & $0.20$ & 4,789 \\
\end{longtable}
\endgroup